\documentclass{article}

\usepackage[final,nonatbib]{nips_2016}

\usepackage[utf8]{inputenc} 
\usepackage[T1]{fontenc}    
\usepackage{hyperref}       
\usepackage{url}            
\usepackage{booktabs}       
\usepackage{amsfonts}       
\usepackage{nicefrac}       
\usepackage{microtype}      

\usepackage{amssymb}

\usepackage{mathtools}

\usepackage{algpseudocode}
\usepackage{algorithm}

\usepackage{graphicx}
\usepackage[lofdepth,lotdepth]{subfig}
\DeclareGraphicsExtensions{.pdf,.png,.jpg}

\title{Learning to Optimize}

\author{
Ke Li \qquad Jitendra Malik \\
Department of Electrical Engineering and Computer Sciences\\
University of California, Berkeley\\
Berkeley, CA 94720\\
United States\\
\texttt{\{ke.li,malik\}@eecs.berkeley.edu} \\
}

\begin{document}

\maketitle

\begin{abstract}
Algorithm design is a laborious process and often requires many iterations of ideation and validation. In this paper, we explore automating algorithm design and present a method to \emph{learn} an optimization algorithm, which we believe to be the first method that can automatically discover a better algorithm. We approach this problem from a reinforcement learning perspective and represent any particular optimization algorithm as a policy. We learn an optimization algorithm using guided policy search and demonstrate that the resulting algorithm outperforms existing hand-engineered algorithms in terms of convergence speed and/or the final objective value. 
\end{abstract}

\section{Introduction}

The current approach to designing algorithms is a laborious process. First, the designer must study the problem and devise an algorithm guided by a mixture of intuition, theoretical and/or empirical insight and general design paradigms. She then needs to analyze the algorithm's performance on prototypical examples and compare it to that of existing algorithms. If the algorithm falls short, she must uncover the underlying cause and find clever ways to overcome the discovered shortcomings. She iterates on this process until she arrives at an algorithm that is superior than existing algorithms. Given the often protracted nature of this process, a natural question to ask is: can we automate it?

In this paper, we focus on automating the design of unconstrained continuous optimization algorithms, which are some of the most powerful and ubiquitous tools used in all areas of science and engineering. Extensive work over the past several decades has yielded many popular methods, like gradient descent, momentum, conjugate gradient and L-BFGS. These algorithms share one commonality: they are all hand-engineered -- that is, the steps of these algorithms are carefully designed by human experts. Just as deep learning has achieved tremendous success by automating feature engineering, automating algorithm design could open the way to similar performance gains. 

We learn a better optimization algorithm by observing its execution. To this end, we formulate the problem as a reinforcement learning problem. Under this framework, any particular optimization algorithm simply corresponds to a policy. We reward optimization algorithms that converge quickly and penalize those that do not. Learning an optimization algorithm then reduces to finding an optimal policy, which can be solved using any reinforcement learning method. To differentiate the algorithm that performs learning from the algorithm that is learned, we will henceforth refer to the former as the ``learning algorithm'' or ``learner'' and the latter as the ``autonomous algorithm'' or ``policy''. We use an off-the-shelf reinforcement learning algorithm known as guided policy search~\cite{levine2014learning}, which has demonstrated success in a variety of robotic control settings~\cite{levine2015end,finn2015learning,levine2015learning,han2015learning}. We show empirically that the autonomous optimization algorithm we learn converges faster and/or finds better optima than existing hand-engineered optimization algorithms. 

\section{Related Work}

Early work has explored the general theme of speeding up learning with accumulation of learning experience. This line of work, known as ``learning to learn'' or ``meta-learning''~\cite{nips1995workshop,vilalta2002perspective,brazdil2008metalearning,thrun2012learning}, considers the problem of devising methods that can take advantage of knowledge learned on other related tasks to train faster, a problem that is today better known as multi-task learning and transfer learning. In contrast, the proposed method can learn to accelerate the training procedure itself, without necessarily requiring any training on related auxiliary tasks. 

A different line of work, known as ``programming by demonstration''~\cite{cypher1993watch}, considers the problem of learning programs from examples of input and output. Several different approaches have been proposed: Liang et al.~\cite{liang2010learning} represents programs explicitly using a formal language, constructs a hierarchical Bayesian prior over programs and performs inference using an MCMC sampling procedure and Graves et al.~\cite{graves2014neural} represents programs implicitly as sequences of memory access operations and trains a recurrent neural net to learn the underlying patterns in the memory access operations. Subsequent work proposes variants of this model that use different primitive memory access operations~\cite{joulin2015inferring}, more expressive operations~\cite{kurach2015neural,yang2016lie} or other non-differentiable operations~\cite{zaremba2015reinforcement,zaremba2015learning}. Others consider building models that permit parallel execution~\cite{kaiser2015neural} or training models with stronger supervision in the form of execution traces~\cite{reed2015neural}. The aim of this line of work is to replicate the behaviour of simple existing algorithms from examples, rather than to learn a new algorithm that is better than existing algorithms. 

There is a rich body of work on hyperparameter optimization, which studies the optimization of hyperparameters used to train a model, such as the learning rate, the momentum decay factor and regularization parameters. Most methods~\cite{hutter2011sequential,bergstra2011algorithms,snoek2012practical,swersky2013multi,feurer2015initializing} rely on sequential model-based Bayesian optimization~\cite{mockus1978application,brochu2010tutorial}, while others adopt a random search approach~\cite{bergstra2012random} or use gradient-based optimization~\cite{bengio2000gradient,domke2012generic,maclaurin2015gradient}. Because each hyperparameter setting corresponds to a particular instantiation of an optimization algorithm, these methods can be viewed as a way to search over different instantiations of the same optimization algorithm. The proposed method, on the other hand, can search over the space of all possible optimization algorithms. In addition, when presented with a new objective function, hyperparameter optimization needs to conduct multiple trials with different hyperparameter settings to find the optimal hyperparameters. In contrast, once training is complete, the autonomous algorithm knows how to choose hyperparameters on-the-fly without needing to try different hyperparameter settings, even when presented with an objective function that it has not seen during training. 

To the best of our knowledge, the proposed method represents the first attempt to learn a better algorithm automatically.

\section{Method}

\subsection{Preliminaries}

In the reinforcement learning setting, the learner is given a choice of actions to take in each time step, which changes the state of the environment in an unknown fashion, and receives feedback based on the consequence of the action. The feedback is typically given in the form of a reward or cost, and the objective of the learner is to choose a sequence of actions based on observations of the current environment that maximizes cumulative reward or minimizes cumulative cost over all time steps. 

A reinforcement learning problem is typically formally represented as an Markov decision process (MDP). We consider a finite-horizon MDP with continuous state and action spaces defined by the tuple $(\mathcal{S},\mathcal{A},p_{0},p,c,\gamma)$, where $\mathcal{S}$ is the set of states, $\mathcal{A}$ is the set of actions, $p_{0}: \mathcal{S} \rightarrow \mathbb{R}^{+}$ is the probability density over initial states, $p: \mathcal{S} \times \mathcal{A} \times \mathcal{S} \rightarrow \mathbb{R}^{+}$ is the transition probability density, that is, the conditional probability density over successor states given the current state and action, $c: \mathcal{S} \rightarrow \mathbb{R}$ is a function that maps state to cost and $\gamma \in (0,1]$ is the discount factor. 

The objective is to learn a stochastic policy $\pi^{*}: \mathcal{S} \times \mathcal{A} \rightarrow \mathbb{R}^{+}$, which is a conditional probability density over actions given the current state, such that the expected cumulative cost is minimized. That is, 
\[
\pi^{*} = \arg\min_{\pi}\mathbb{E}_{s_{0},a_{0},s_{1},\ldots,s_{T}}\left[\sum_{t=0}^{T}\gamma^{t}c(s_{t})\right], 
\]
where the expectation is taken with respect to the joint distribution over the sequence of states and actions, often referred to as a trajectory, which has the density
\[
q\left(s_{0},a_{0},s_{1},\ldots,s_{T}\right)=p_{0}\left(s_{0}\right)\prod_{t=0}^{T-1}\pi\left(\left.a_{t}\right|s_{t}\right)p\left(\left.s_{t+1}\right|s_{t},a_{t}\right). 
\]
This problem of finding the cost-minimizing policy is known as the policy search problem. To enable generalization to unseen states, the policy is typically parameterized and minimization is performed over representable policies. Solving this problem exactly is intractable in all but selected special cases. Therefore, policy search methods generally tackle this problem by solving it approximately. 

In many practical settings, $p$, which characterizes the dynamics, is unknown and must therefore be estimated. Additionally, because it is often equally important to minimize cost at earlier and later time steps, we will henceforth focus on the undiscounted setting, i.e. the setting where $\gamma = 1$. 

Guided policy search~\cite{levine2014learning} is a method for performing policy search in continuous state and action spaces under possibly unknown dynamics. It works by alternating between computing a target distribution over trajectories that is encouraged to minimize cost and agree with the current policy and learning parameters of the policy in a standard supervised fashion so that sample trajectories from executing the policy are close to sample trajectories drawn from the target distribution. The target trajectory distribution is computed by iteratively fitting local time-varying linear and quadratic approximations to the (estimated) dynamics and cost respectively and optimizing over a restricted class of linear-Gaussian policies subject to a trust region constraint, which can be solved efficiently in closed form using a dynamic programming algorithm known as linear-quadratic-Gaussian (LQG). We refer interested readers to \cite{levine2014learning} for details. 

\subsection{Formulation}

Consider the general structure of an algorithm for unconstrained continuous optimization, which is outlined in Algorithm \ref{alg:opt_alg_structure}. Starting from a random location in the domain of the objective function, the algorithm iteratively updates the current location by a step vector computed from some functional $\pi$ of the objective function, the current location and past locations. 

\begin{algorithm}
\caption{General structure of optimization algorithms}
\label{alg:opt_alg_structure}
\begin{algorithmic}
\Require Objective function $f$
\State $x^{(0)} \gets $ random point in the domain of $f$
\For{$i = 1,2,\ldots$}
    \State $\Delta x \gets \pi (f, \{x^{(0)},\ldots,x^{(i-1)}\})$
    \If{stopping condition is met}
        \State \Return{$x^{(i-1)}$}
    \EndIf
    \State $x^{(i)} \gets x^{(i-1)} + \Delta x$
\EndFor
\end{algorithmic}
\end{algorithm}

This framework subsumes all existing optimization algorithms. Different optimization algorithms differ in the choice of $\pi$. First-order methods use a $\pi$ that depends only on the gradient of the objective function, whereas second-order methods use a $\pi$ that depends on both the gradient and the Hessian of the objective function. In particular, the following choice of $\pi$ yields the gradient descent method:
\[
\pi (f, \{x^{(0)},\ldots,x^{(i-1)}\}) = -\gamma \nabla f (x^{(i-1)}), 
\]
where $\gamma$ denotes the step size or learning rate. Similarly, the following choice of $\pi$ yields the gradient descent method with momentum:
\[
\pi (f, \{x^{(0)},\ldots,x^{(i-1)}\}) = -\gamma \left( \sum_{j=0}^{i-1}\alpha^{i-1-j}\nabla f(x^{(j)}) \right), 
\]
where $\gamma$ again denotes the step size and $\alpha$ denotes the momentum decay factor. 

Therefore, if we can learn $\pi$, we will be able to learn an optimization algorithm. Since it is difficult to model general functionals, in practice, we restrict the dependence of $\pi$ on the objective function $f$ to objective values and gradients evaluated at current and past locations. Hence, $\pi$ can be simply modelled as a function from the objective values and gradients along the trajectory taken by the optimizer so far to the next step vector. 

We observe that the execution of an optimization algorithm can be viewed as the execution of a fixed policy in an MDP: the state consists of the current location and the objective values and gradients evaluated at the current and past locations, the action is the step vector that is used to update the current location, and the transition probability is partially characterized by the location update formula, $x^{(i)} \gets x^{(i-1)} + \Delta x$. The policy that is executed corresponds precisely to the choice of $\pi$ used by the optimization algorithm. For this reason, we will also use $\pi$ to denote the policy at hand. Under this formulation, searching over policies corresponds to searching over all possible first-order optimization algorithms. 

We can use reinforcement learning to learn the policy $\pi$. To do so, we need to define the cost function, which should penalize policies that exhibit undesirable behaviours during their execution. Since the performance metric of interest for optimization algorithms is the speed of convergence, the cost function should penalize policies that converge slowly. To this end, assuming the goal is to minimize the objective function, we define cost at a state to be the objective value at the current location. This encourages the policy to reach the minimum of the objective function as quickly as possible. 

Since the policy $\pi$ may be stochastic in general, we model each dimension of the action conditional on the state as an independent Gaussian whose mean is given by a regression model and variance is some learned constant. We choose to parameterize the mean of $\pi$ using a neural net, due to its appealing properties as a universal function approximator and strong empirical performance in a variety of applications. We use guided policy search to learn the parameters of the policy. 

We use a training set consisting of different randomly generated objective functions. We evaluate the resulting autonomous algorithm on different objective functions drawn from the same distribution. 

\subsection{Discussion}

An autonomous optimization algorithm offers several advantages over hand-engineered algorithms. First, an autonomous optimizer is trained on real algorithm execution data, whereas hand-engineered optimizers are typically derived by analyzing objective functions with properties that may or may not be satisfied by objective functions that arise in practice. Hence, an autonomous optimizer minimizes the amount of a priori assumptions made about objective functions and can instead take full advantage of the information about the actual objective functions of interest. Second, an autonomous optimizer has no hyperparameters that need to be tuned by the user. Instead of just computing a step direction which must then be combined with a user-specified step size, an autonomous optimizer predicts the step direction and size jointly. This allows the autonomous optimizer to dynamically adjust the step size based on the information it has acquired about the objective function while performing the optimization. Finally, when an autonomous optimizer is trained on a particular class of objective functions, it may be able to discover hidden structure in the geometry of the class of objective functions. At test time, it can then exploit this knowledge to perform optimization faster. 

\subsection{Implementation Details}

We store the current location, previous gradients and improvements in the objective value from previous iterations in the state. We keep track of only the information pertaining to the previous $H$ time steps and use $H = 25$ in our experiments. More specifically, the dimensions of the state space encode the following information:
\begin{itemize}
\item Current location in the domain
\item Change in the objective value at the current location relative to the objective value at the $i^{\mathrm{th}}$ most recent location for all $i \in \{2,\ldots,H+1\}$
\item Gradient of the objective function evaluated at the $i^{\mathrm{th}}$ most recent location for all $i \in \{2,\ldots,H+1\}$
\end{itemize}

Initially, we set the dimensions corresponding to historical information to zero. The current location is only used to compute the cost; because the policy should not depend on the absolute coordinates of the current location, we exclude it from the input that is fed into the neural net. 

We use a small neural net to model the policy. Its architecture consists of a single hidden layer with 50 hidden units. Softplus activation units are used in the hidden layer and linear activation units are used in the output layer. The training objective imposed by guided policy search takes the form of the squared Mahalanobis distance between mean predicted and target actions along with other terms dependent on the variance of the policy. We also regularize the entropy of the policy to encourage deterministic actions conditioned on the state. The coefficient on the regularizer increases gradually in later iterations of guided policy search. We initialize the weights of the neural net randomly and do not regularize the magnitude of weights. 

Initially, we set the target trajectory distribution so that the mean action given state at each time step matches the step vector used by the gradient descent method with momentum. We choose the best settings of the step size and momentum decay factor for each objective function in the training set by performing a grid search over hyperparameters and running noiseless gradient descent with momentum for each hyperparameter setting. 

For training, we sample 20 trajectories with a length of 40 time steps for each objective function in the training set. After each iteration of guided policy search, we sample new trajectories from the new distribution and discard the trajectories from the preceding iteration. 

\section{Experiments}

We learn autonomous optimization algorithms for various convex and non-convex classes of objective functions that correspond to loss functions for different machine learning models. We first learn an autonomous optimizer for logistic regression, which induces a convex loss function. We then learn an autonomous optimizer for robust linear regression using the Geman-McClure M-estimator, whose loss function is non-convex. Finally, we learn an autonomous optimizer for a two-layer neural net classifier with ReLU activation units, whose error surface has even more complex geometry. 

\subsection{Logistic Regression}

We consider a logistic regression model with an $\ell_{2}$ regularizer on the weight vector. Training the model requires optimizing the following objective:
\[
\min_{\mathbf{w},b}-\frac{1}{n}\sum_{i=1}^{n}y_{i}\log\sigma\left(\mathbf{w}^{T}\mathbf{x}_{i}+b\right)+(1-y_{i})\log\left(1-\sigma\left(\mathbf{w}^{T}\mathbf{x}_{i}+b\right)\right)+\frac{\lambda}{2}\left\Vert \mathbf{w}\right\Vert _{2}^{2},
\]
where $\mathbf{w}\in\mathbb{R}^{d}$ and $b\in\mathbb{R}$ denote the weight vector and bias respectively, $\mathbf{x}_{i}\in\mathbb{R}^{d}$ and $y_{i}\in\{0,1\}$ denote the feature vector and label of the $i^{\mathrm{th}}$ instance, $\lambda$ denotes the coefficient on the regularizer and $\sigma(z) \coloneqq \frac{1}{1+e^{-z}}$. For our experiments, we choose $\lambda = 0.0005$ and $d = 3$. This objective is convex in $\mathbf{w}$ and $b$. 

We train an autonomous algorithm that learns to optimize objectives of this form. The training set consists of examples of such objective functions whose free variables, which in this case are $\mathbf{x}_{i}$ and $y_{i}$, are all assigned concrete values. Hence, each objective function in the training set corresponds to a logistic regression problem on a different dataset. 

To construct the training set, we randomly generate a dataset of 100 instances for each function in the training set. The instances are drawn randomly from two multivariate Gaussians with random means and covariances, with half drawn from each. Instances from the same Gaussian are assigned the same label and instances from different Gaussians are assigned different labels. 

We train the autonomous algorithm on a set of $90$ objective functions. We evaluate it on a test set of $100$ random objective functions generated using the same procedure and compare to popular hand-engineered algorithms, such as gradient descent, momentum, conjugate gradient and L-BFGS. All baselines are run with the best hyperparameter settings tuned on the training set. 

\begin{figure}
    \centering
    \subfloat[]{
        \includegraphics[width=0.33\textwidth]{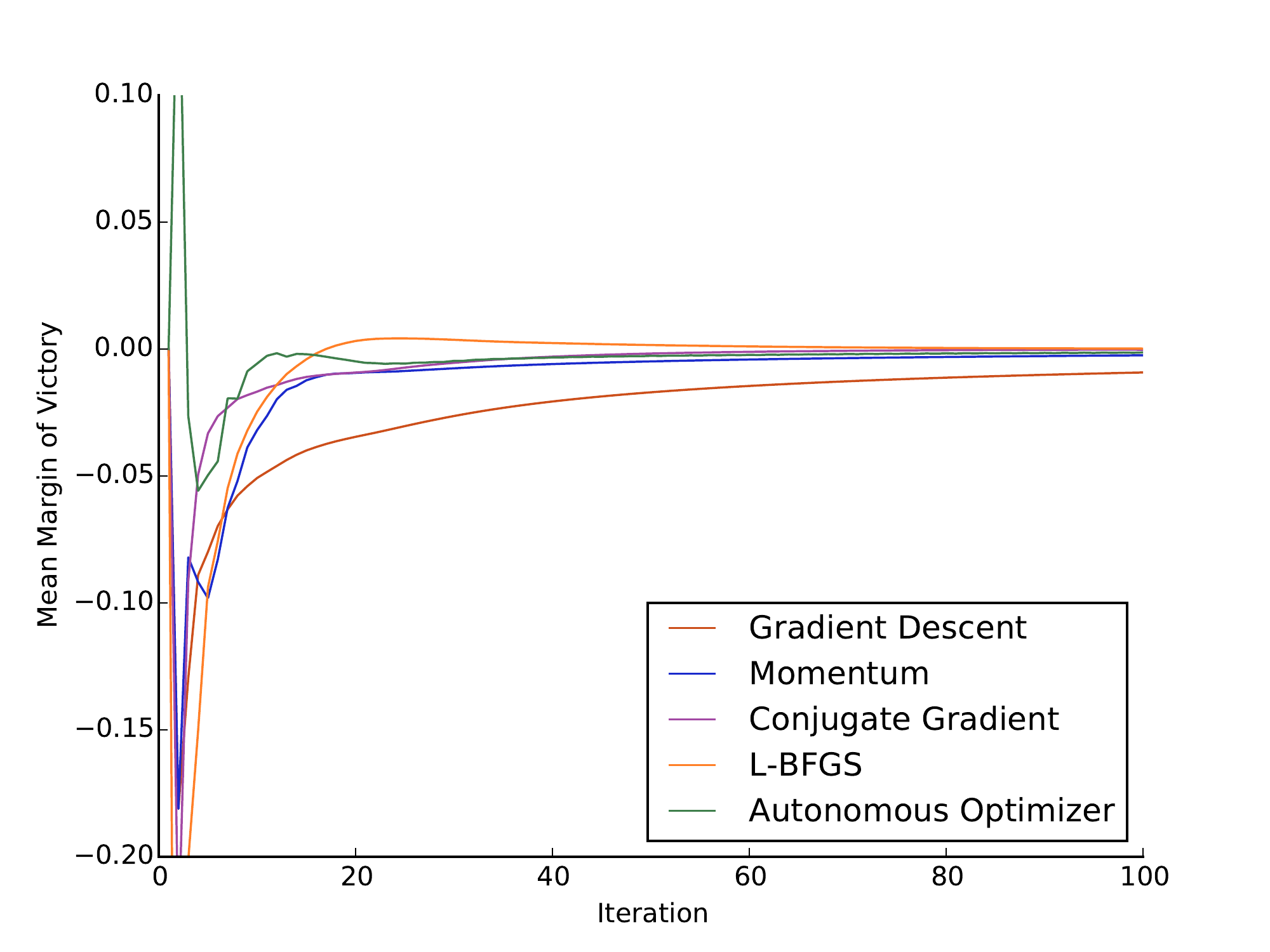}
        \label{fig:logisticreg_mean}
    }
    \subfloat[]{
        \includegraphics[width=0.33\textwidth]{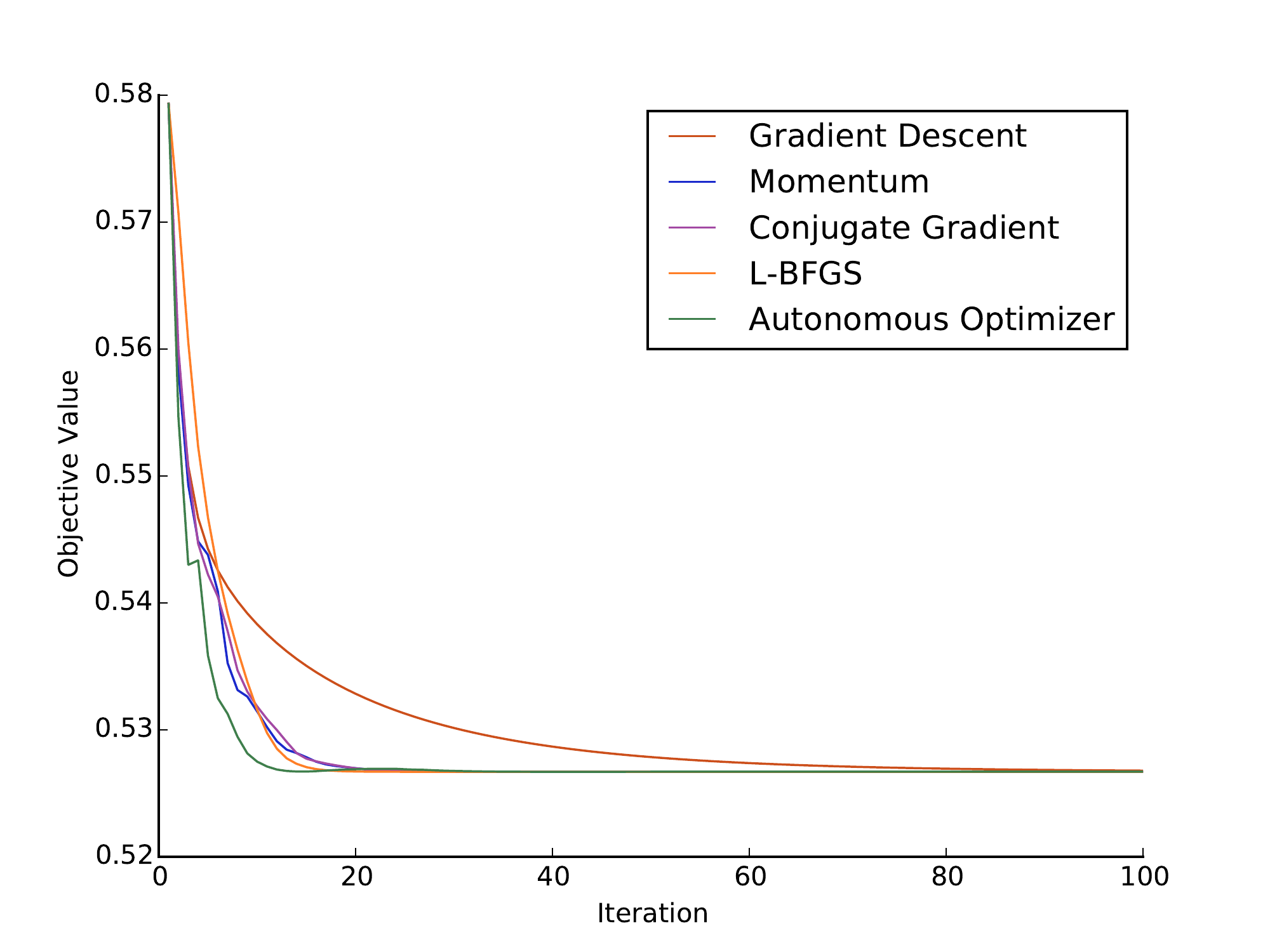}
        \label{fig:logisticreg_a}
    }
    \subfloat[]{
        \includegraphics[width=0.33\textwidth]{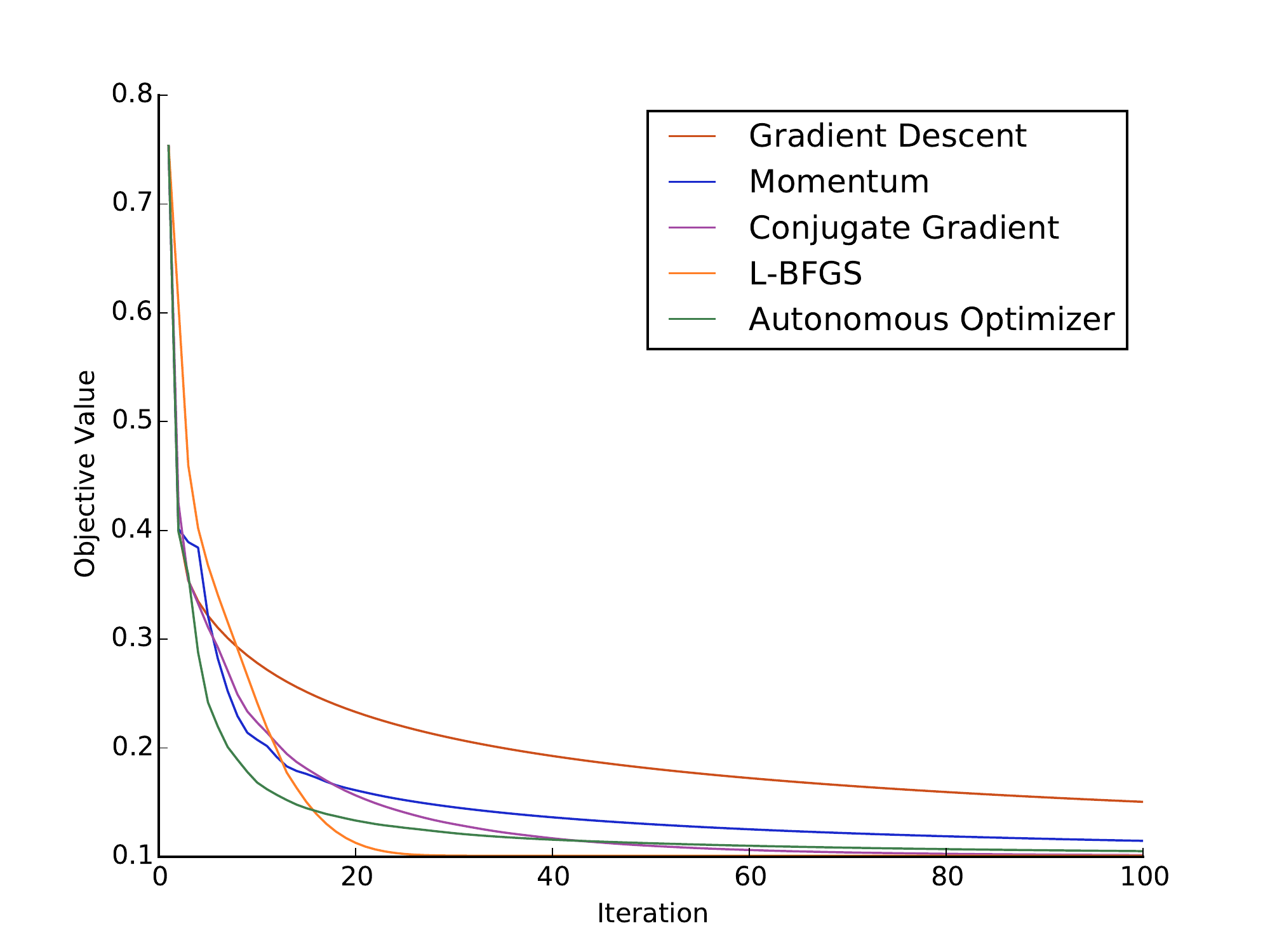}
        \label{fig:logisticreg_b}
    }
    \caption{\label{fig:logisticreg_results}(a) Mean margin of victory of each algorithm for optimizing the logistic regression loss. Higher margin of victory indicates better performance. (b-c) Objective values achieved by each algorithm on two objective functions from the test set. Lower objective values indicate better performance. Best viewed in colour.}
\end{figure}

For each algorithm and objective function in the test set, we compute the difference between the objective value achieved by a given algorithm and that achieved by the best of the competing algorithms at every iteration, a quantity we will refer to as ``the margin of victory''. This quantity is positive when the current algorithm is better than all other algorithms and negative otherwise. In Figure \ref{fig:logisticreg_mean}, we plot the mean margin of victory of each algorithm at each iteration averaged over all objective functions in the test set. We find that conjugate gradient and L-BFGS diverge or oscillate in rare cases (on 6\% of the objective functions in the test set), even though the autonomous algorithm, gradient descent and momentum do not. To reflect performance of these baselines in the majority of cases, we exclude the offending objective functions when computing the mean margin of victory. 

As shown, the autonomous algorithm outperforms gradient descent, momentum and conjugate gradient at almost every iteration. The margin of victory of the autonomous algorithm is quite high in early iterations, indicating that the autonomous algorithm converges much faster than other algorithms. It is interesting to note that despite having seen only trajectories of length 40 at training time, the autonomous algorithm is able to generalize to much longer time horizons at test time. L-BFGS converges to slightly better optima than the autonomous algorithm and the momentum method. This is not surprising, as the objective functions are convex and L-BFGS is known to be a very good optimizer for convex optimization problems. 

We show the performance of each algorithm on two objective functions from the test set in Figures \ref{fig:logisticreg_a} and \ref{fig:logisticreg_b}. In Figure \ref{fig:logisticreg_a}, the autonomous algorithm converges faster than all other algorithms. In Figure \ref{fig:logisticreg_b}, the autonomous algorithm initially converges faster than all other algorithms but is later overtaken by L-BFGS, while remaining faster than all other optimizers. However, it eventually achieves the same objective value as L-BFGS, while the objective values achieved by gradient descent and momentum remain much higher. 

\subsection{Robust Linear Regression}

Next, we consider the problem of linear regression using a robust loss function. One way to ensure robustness is to use an M-estimator for parameter estimation. A popular choice is the Geman-McClure estimator, which induces the following objective:
\[
\min_{\mathbf{w},b}\frac{1}{n}\sum_{i=1}^{n}\frac{\left(y_{i}-\mathbf{w}^{T}\mathbf{x}_{i}-b\right)^{2}}{c^{2}+\left(y_{i}-\mathbf{w}^{T}\mathbf{x}_{i}-b\right)^{2}},
\]
where $\mathbf{w}\in\mathbb{R}^{d}$ and $b\in\mathbb{R}$ denote the weight vector and bias respectively, $\mathbf{x}_{i}\in\mathbb{R}^{d}$ and $y_{i}\in\mathbb{R}$ denote the feature vector and label of the $i^{\mathrm{th}}$ instance and $c\in\mathbb{R}$ is a constant that modulates the shape of the loss function. For our experiments, we use $c = 1$ and $d = 3$. This loss function is not convex in either $\mathbf{w}$ or $b$. 

As with the preceding section, each objective function in the training set is a function of the above form with realized values for $\mathbf{x}_{i}$ and $y_{i}$. The dataset for each objective function is generated by drawing 25 random samples from each one of four multivariate Gaussians, each of which has a random mean and the identity covariance matrix. For all points drawn from the same Gaussian, their labels are generated by projecting them along the same random vector, adding the same randomly generated bias and perturbing them with i.i.d. Gaussian noise. 

\begin{figure}
    \centering
    \subfloat[]{
        \includegraphics[width=0.33\textwidth]{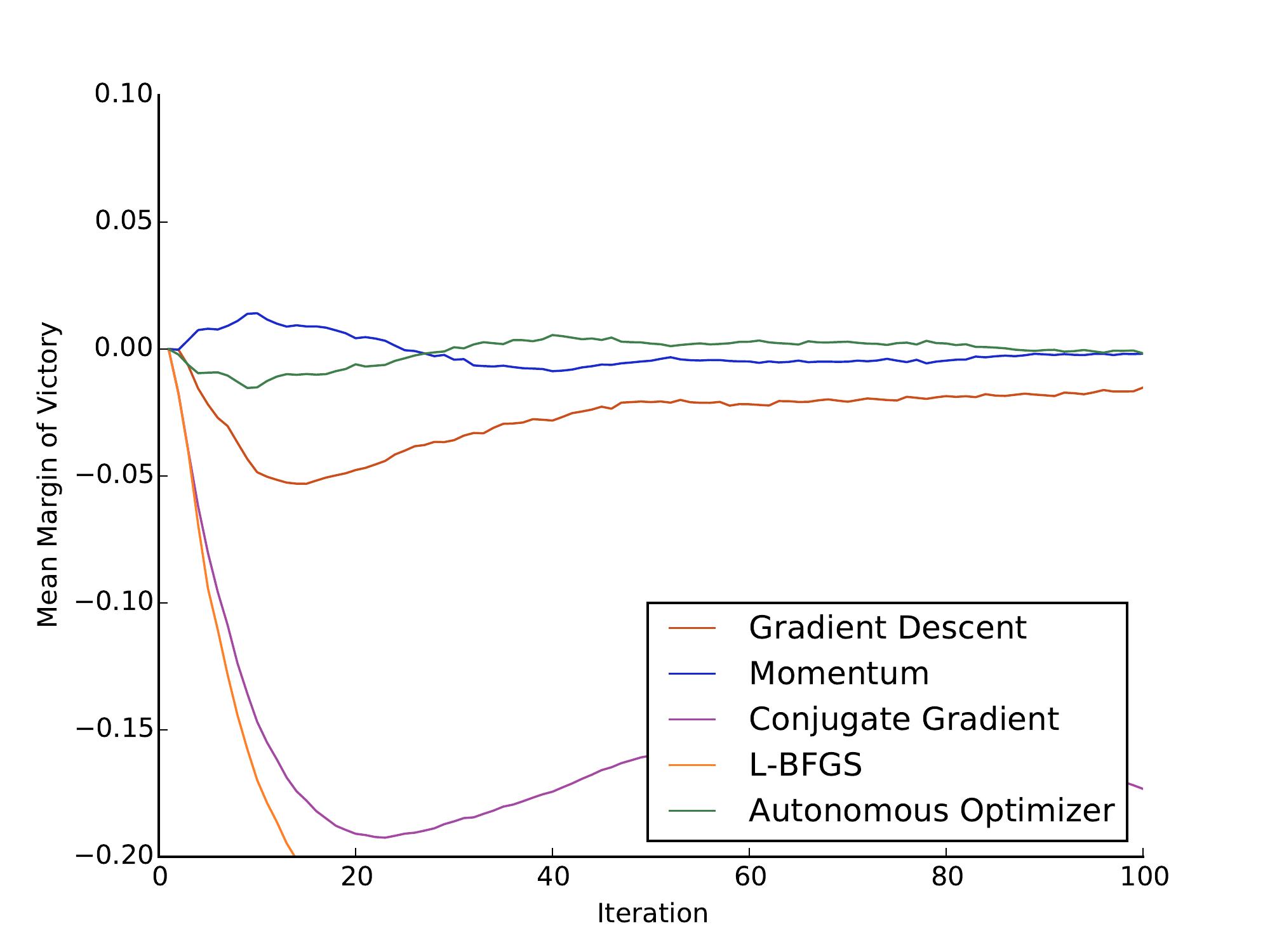}
        \label{fig:robustreg_mean}
    }
    \subfloat[]{
        \includegraphics[width=0.33\textwidth]{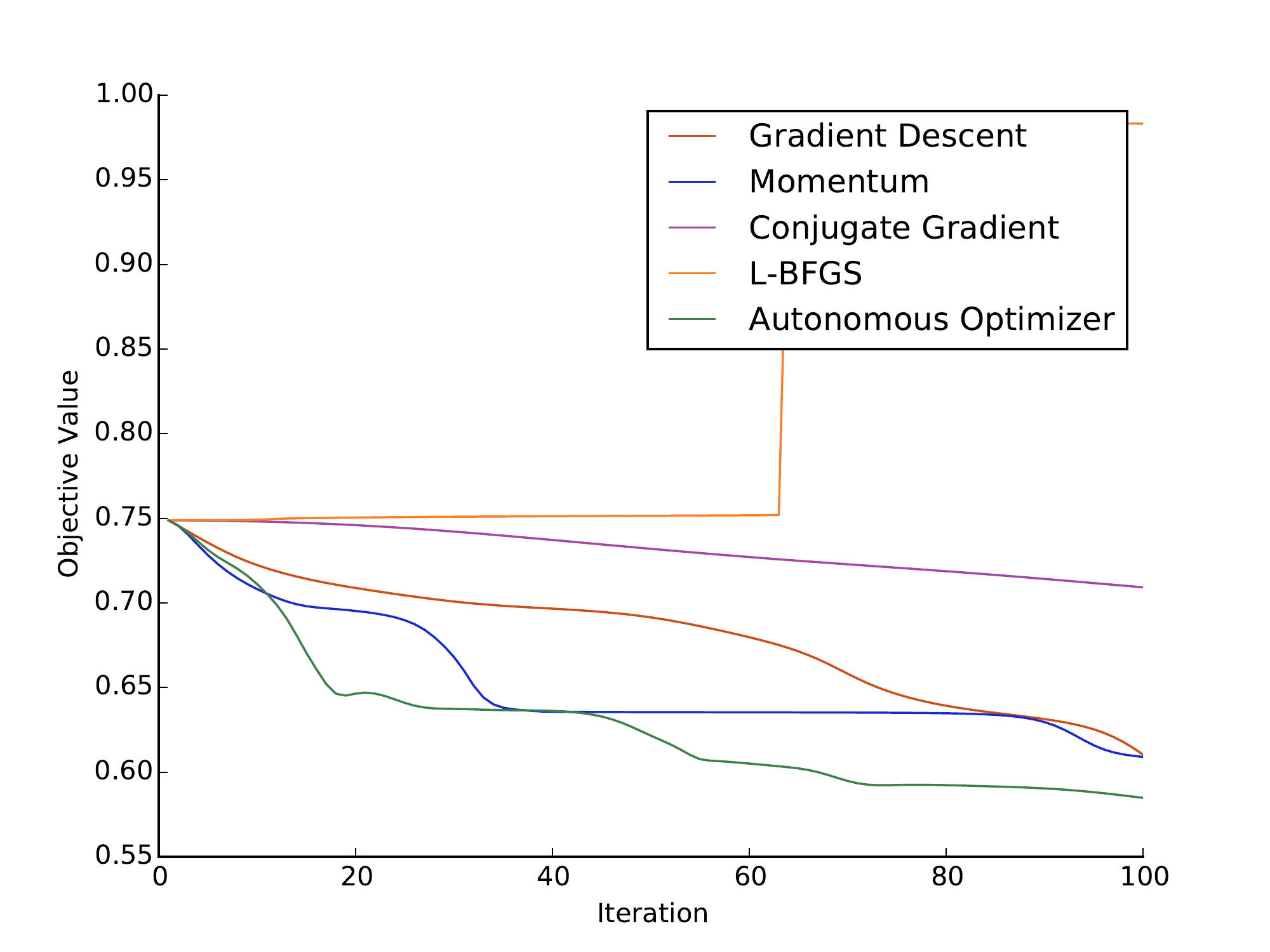}
        \label{fig:robustreg_a}
    }
    \subfloat[]{
        \includegraphics[width=0.33\textwidth]{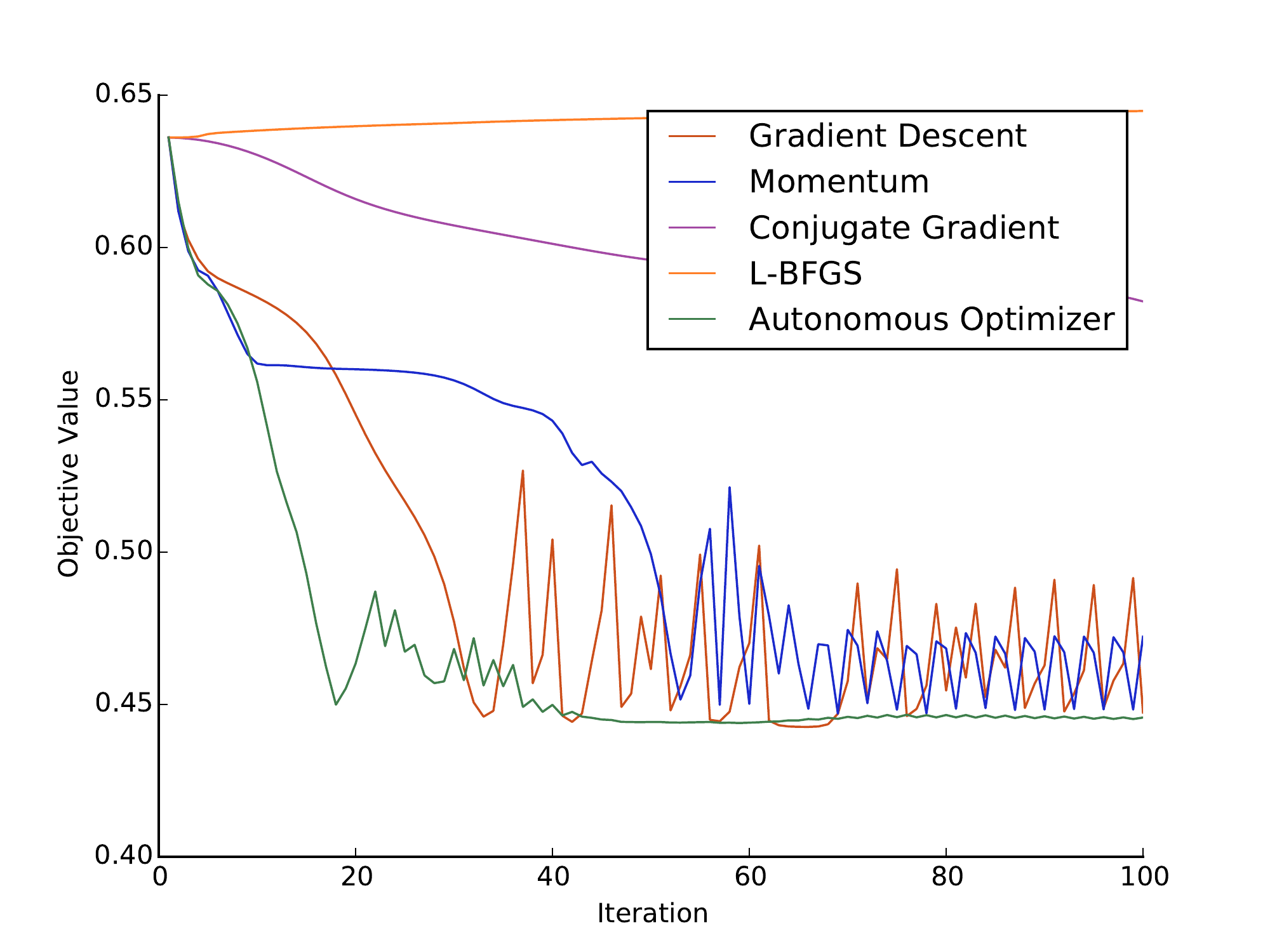}
        \label{fig:robustreg_b}
    }
    \caption{\label{fig:robustreg_results}(a) Mean margin of victory of each algorithm for optimizing the robust linear regression loss. Higher margin of victory indicates better performance. (b-c) Objective values achieved by each algorithm on two objective functions from the test set. Lower objective values indicate better performance. Best viewed in colour.}
\end{figure}

The autonomous algorithm is trained on a set of 120 objective functions. We evaluate it on 100 randomly generated objective functions using the same metric as above. As shown in Figure \ref{fig:robustreg_mean}, the autonomous algorithm outperforms all hand-engineered algorithms except at early iterations. While it dominates gradient descent, conjugate gradient and L-BFGS at all times, it does not make progress as quickly as the momentum method initially. However, after around 30 iterations, it is able to close the gap and surpass the momentum method. On this optimization problem, both conjugate gradient and L-BFGS diverge quickly. Interestingly, unlike in the previous experiment, L-BFGS no longer performs well, which could be caused by non-convexity of the objective functions. 

Figures \ref{fig:robustreg_a} and \ref{fig:robustreg_b} show performance on objective functions from the test set. In Figure \ref{fig:robustreg_a}, the autonomous optimizer not only converges the fastest, but also reaches a better optimum than all other algorithms. In Figure \ref{fig:robustreg_b}, the autonomous algorithm converges the fastest and is able to avoid most of the oscillations that hamper gradient descent and momentum after reaching the optimum. 

\subsection{Neural Net Classifier}

Finally, we train an autonomous algorithm to train a small neural net classifier. We consider a two-layer neural net with ReLU activation on the hidden units and softmax activation on the output units. We use the cross-entropy loss combined with $\ell_{2}$ regularization on the weights. To train the model, we need to optimize the following objective:
\[
\min_{W,U,\mathbf{b},\mathbf{c}}-\frac{1}{n}\sum_{i=1}^{n}\log\left(\frac{\exp\left(\left(U\max\left(W\mathbf{x}_{i}+\mathbf{b},0\right)+\mathbf{c}\right)_{y_{i}}\right)}{\sum_{j}\exp\left(\left(U\max\left(W\mathbf{x}_{i}+\mathbf{b},0\right)+\mathbf{c}\right)_{j}\right)}\right)+\frac{\lambda}{2}\left\Vert W\right\Vert _{F}^{2}+\frac{\lambda}{2}\left\Vert U\right\Vert _{F}^{2},
\]
where $W\in\mathbb{R}^{h\times d},b\in\mathbb{R}^{h},U\in\mathbb{R}^{p\times h},c\in\mathbb{R}^{p}$ denote the first-layer and second-layer weights and biases, $\mathbf{x}_{i}\in\mathbb{R}^{d}$ and $y_{i}\in\{1,\ldots,p\}$ denote the input and target class label of the $i^{\mathrm{th}}$ instance, $\lambda$ denotes the coefficient on regularizers and $\left(\mathbf{v}\right)_{j}$ denotes the $j^{\mathrm{th}}$ component of $\mathbf{v}$. For our experiments, we use $\lambda = 0.0005$ and $d = h = p = 2$. The error surface is known to have complex geometry and multiple local optima, making this a challenging optimization problem. 

The training set consists of 80 objective functions, each of which corresponds to the objective for training a neural net on a different dataset. Each dataset is generated by generating four multivariate Gaussians with random means and covariances and sampling 25 points from each. The points from the same Gaussian are assigned the same random label of either 0 or 1. We make sure not all of the points in the dataset are assigned the same label. 

\begin{figure}
    \centering
    \subfloat[]{
        \includegraphics[width=0.33\textwidth]{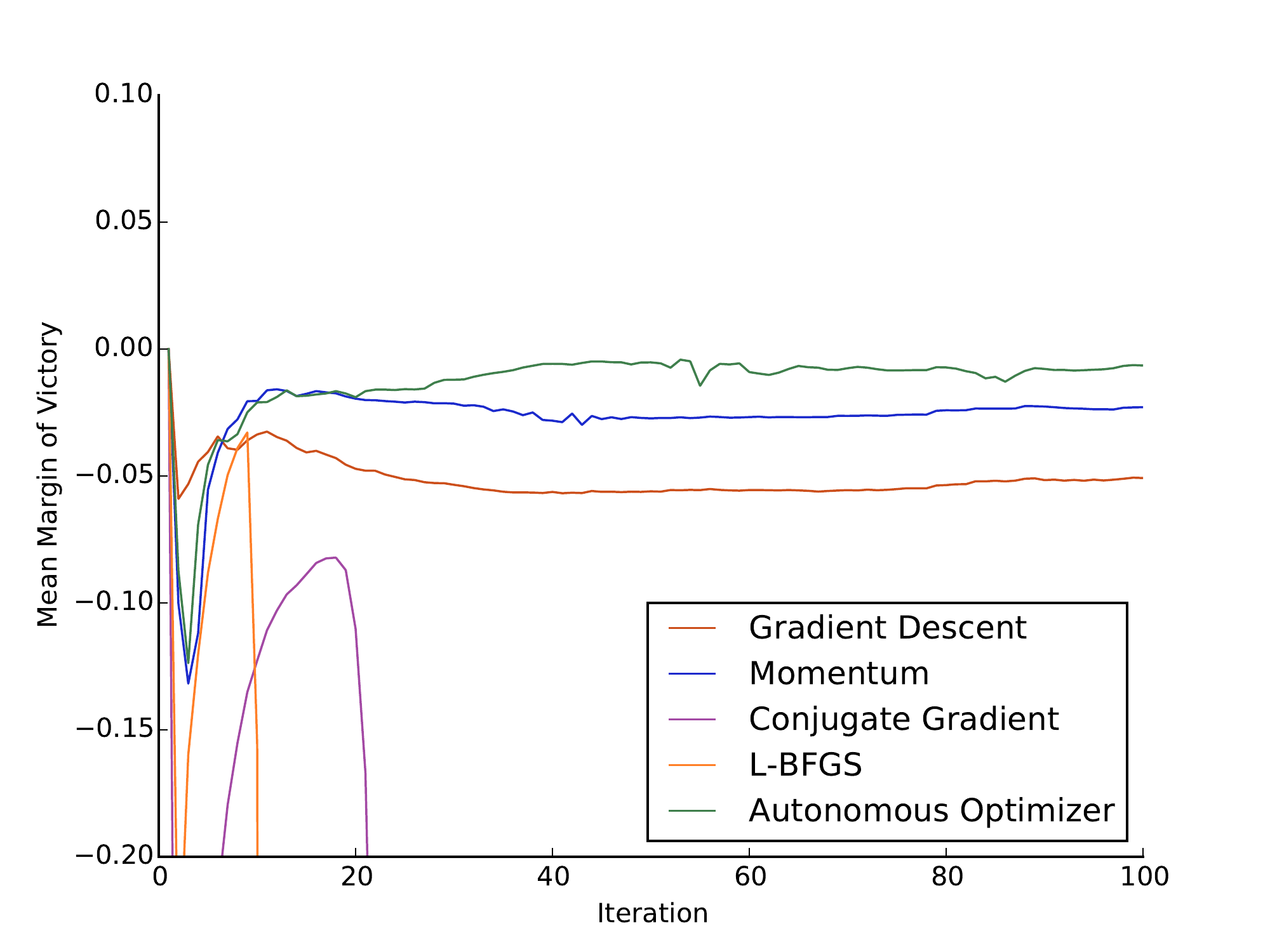}
        \label{fig:neuralnet_mean}
    }
    \subfloat[]{
        \includegraphics[width=0.33\textwidth]{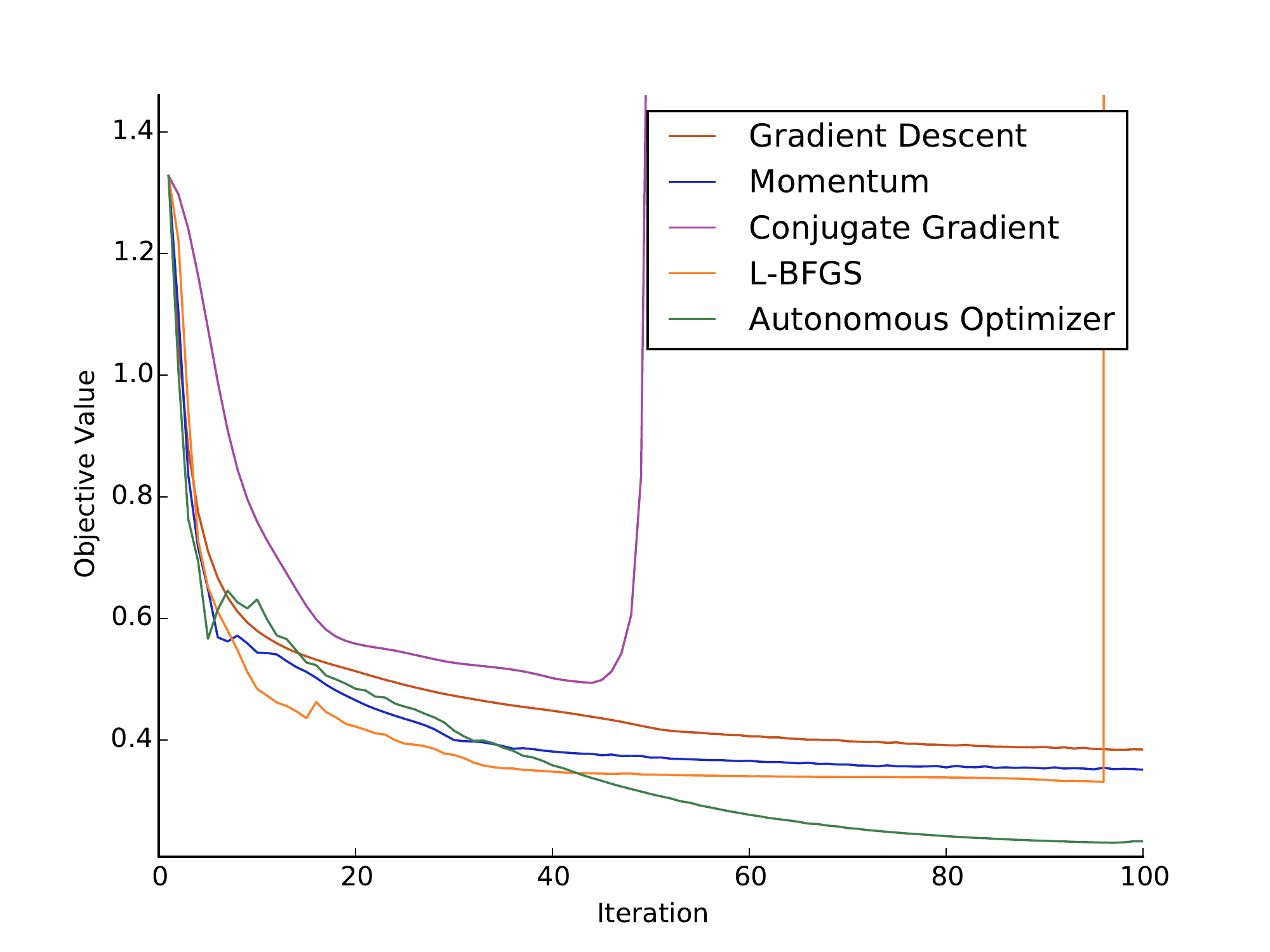}
        \label{fig:neuralnet_a}
    }
    \subfloat[]{
        \includegraphics[width=0.33\textwidth]{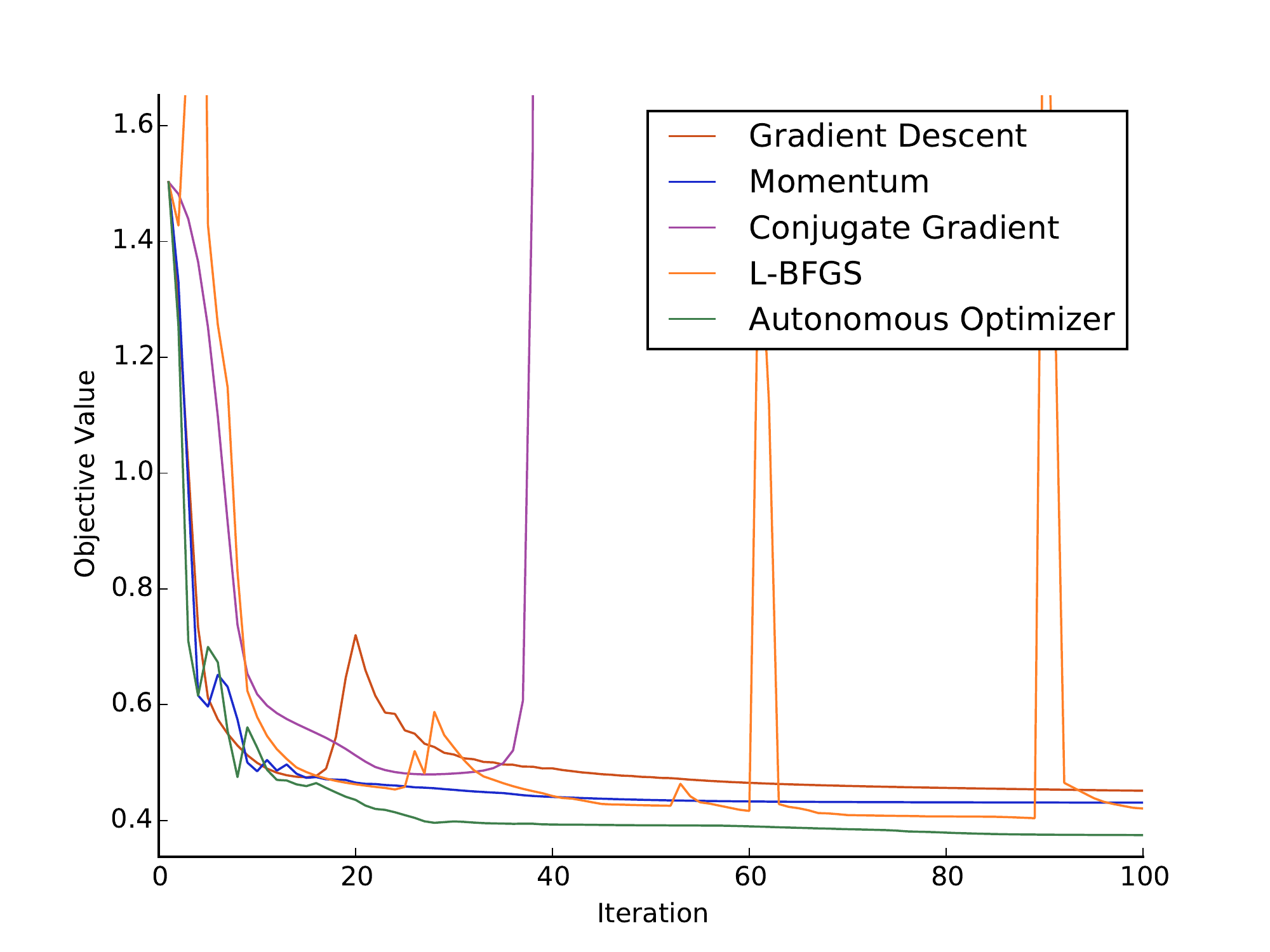}
        \label{fig:neuralnet_b}
    }
    \caption{\label{fig:neuralnet_results}
    (a) Mean margin of victory of each algorithm for training neural net classifiers. Higher margin of victory indicates better performance. (b-c) Objective values achieved by each algorithm on two objective functions from the test set. Lower objective values indicate better performance. Best viewed in colour.}
\end{figure}

We evaluate the autonomous algorithm in the same manner as above. As shown in Figure \ref{fig:neuralnet_mean}, the autonomous algorithm significantly outperforms all other algorithms. In particular, as evidenced by the sizeable and sustained gap between margin of victory of the autonomous optimizer and the momentum method, the autonomous optimizer is able to reach much better optima and is less prone to getting trapped in local optima compared to other methods. This gap is also larger compared to that exhibited in previous sections, suggesting that hand-engineered algorithms are more sub-optimal on challenging optimization problems and so the potential for improvement from learning the algorithm is greater in such settings. Due to non-convexity, conjugate gradient and L-BFGS often diverge. 

Performance on examples of objective functions from the test set is shown in Figures \ref{fig:neuralnet_a} and \ref{fig:neuralnet_b}. As shown, the autonomous optimizer is able to reach better optima than all other methods and largely avoids oscillations that other methods suffer from. 

\section{Conclusion}

We presented a method for learning a better optimization algorithm. We formulated this as a reinforcement learning problem, in which any optimization algorithm can be represented as a policy. Learning an optimization algorithm then reduces to find the optimal policy. We used guided policy search for this purpose and trained autonomous optimizers for different classes of convex and non-convex objective functions. We demonstrated that the autonomous optimizer converges faster and/or reaches better optima than hand-engineered optimizers. We hope autonomous optimizers learned using the proposed approach can be used to solve various common classes of optimization problems more quickly and help accelerate the pace of innovation in science and engineering. 

\small
\bibliographystyle{plain}
\bibliography{lto}

\end{document}